\documentclass[]{article}
\usepackage{spconf}
\ninept 

\usepackage[utf8]{inputenc} 
\usepackage[T1]{fontenc}    
\usepackage{hyperref}       
\usepackage{url}            
\usepackage{booktabs}       
\usepackage{amsfonts}       
\usepackage{nicefrac}       
\usepackage{microtype}      
\usepackage{siunitx}
\usepackage{enumerate}
\usepackage{float}
\usepackage{caption}

\usepackage{lipsum}

\hypersetup{allcolors=blue,colorlinks=true}

\usepackage{graphicx}
\usepackage{amsmath}
\usepackage{amsthm}
\usepackage{amssymb}
\usepackage{amsfonts}
\usepackage{mathrsfs}
\usepackage{multirow}
\usepackage{algorithm}
\usepackage{subfigure}
\usepackage{thm-restate}
\usepackage{mathtools}
\usepackage{comment}
\usepackage{multicol}
\usepackage{bm}

\newcommand{\R}{\mathbb{R}}

\newcommand{\bPsi}{\bm{\Psi}}
\newcommand{\bD}{\bm{D}}

\newcommand{\bI}{\bm{I}}

\newcommand{\bU}{\bm{U}}
\newcommand{\bV}{\bm{V}}

\newcommand{\bX}{\bm{X}}
\newcommand{\bY}{\bm{Y}}

\newcommand{\bomega}{\bm{\omega}}

\newcommand{\bw}{\bm{w}}
\newcommand{\bx}{\bm{x}}

\newcommand{\bZ}{\bm{Z}}
\newcommand{\bzeta}{\bm{\zeta}}

\usepackage{bbm}

\newcommand{\bzero}{\mathbf{0}}

\DeclareMathOperator*{\argmin}{arg\,min}

\DeclareMathOperator{\prox}{\textsf{prox}}

\usepackage{pifont}

\newtheorem{theorem}{Theorem}

\newcommand{\pp}{\phantom{+}}

\usepackage[natbib=true,
            bibstyle=ieee,
            citestyle=numeric-comp,
            uniquename=false,
            uniquelist=false,
            maxcitenames=2,
            mincitenames=1,
            mincrossrefs=99]{biblatex}
\renewbibmacro{in:}{\ifentrytype{article}{}{\printtext{\bibstring{in}\intitlepunct}}}
\DefineBibliographyStrings{english}{url = URL}

\AtEveryBibitem{%
  \clearlist{publisher}%
  \clearfield{number}%
  \clearfield{pagetotal}%
  \clearfield{series}%
  \clearfield{edition}%
  \clearfield{isbn}%
  \clearname{editor}%
  \clearlist{location}%
  \clearfield{eprint}
}

\title{Simultaneous Grouping and Denoising via Sparse Convex Wavelet Clustering}
\name{Michael Weylandt$^{\dagger}$, T. Mitchell Roddenberry$^{\star}$, and Genevera I. Allen$^{\star}$
\address{$^\dagger$University of Florida Informatics Institute, Gainesville, FL USA\\
$^\star$Department of Electrical and Computer Engineering, Rice University, Houston, TX USA\\
\href{mailto:michael.weylandt@ufl.edu}{michael.weylandt@ufl.edu} \quad \href{mailto:mitch@rice.edu}{mitch@rice.edu} \quad \href{mailto:gallen@rice.edu}{gallen@rice.edu}}
\thanks{MW's research is supported by an appointment to the Intelligence Community Postdoctoral Research Fellowship Program at the University of Florida Informatics Institute, administered by Oak Ridge Institute for Science and Education through an interagency agreement between the U.S. Department of Energy and the Office of the Director of National Intelligence. GIA is also affiliated with the Departments of Statistics and of Computer Science at Rice University and with the Neurological Research Institute at Baylor College of Medicine.  GIA acknowledges support from NSF DMS-1554821, NSF NeuroNex-1707400, and NIH 1R01GM140468.  The authors also thank Minjie Wang for helpful discussions on this work.}}

\begin{document}
\begin{refsection}
\maketitle

\begin{abstract}
Clustering is a ubiquitous problem in data science and signal processing.  In many applications where we observe noisy signals, it is common practice to first denoise the data, perhaps using wavelet denoising, and then to apply a clustering algorithm.  In this paper, we develop a sparse convex wavelet clustering approach that simultaneously denoises and discovers groups.  Our approach utilizes convex fusion penalties to achieve agglomeration and group-sparse penalties to denoise through sparsity in the wavelet domain.  In contrast to common practice which denoises then clusters, our method is a unified, convex approach that performs both simultaneously.  Our method yields denoised (wavelet-sparse) cluster centroids that both improve interpretability and data compression.  We demonstrate our method on synthetic examples and in an application to NMR spectroscopy.
\end{abstract}


\section{Introduction}

Clustering seeks to find latent groupings in large and often noisy data sets.  Traditional clustering approaches, such as $K$-means, are known to perform poorly with high-dimensional and noisy signals commonly found in applications such as medical imaging, spectroscopy, and genomics.  In such situations, it is common to first denoise the data, say using wavelet denoising, and then apply clustering techniques to discover groups~\cite{Misti:2007}. This sort of greedy two-step procedure may be undesirable mathematically, as it achieves a local solution to the overarching goal, and practically, as it yields cluster centroids which are not themselves denoised.  In this paper, we seek to discover clusters whose centroids are denoised, yielding more interpretable clustering results.  We propose to achieve this via \emph{sparse convex wavelet clustering}, extending recent work in convex clustering to the wavelet domain in order to yield wavelet-sparse cluster centroids in a unified, convex, and mathematically appealing manner. 

\begin{figure*}
\centering
\includegraphics[height=2.5in]{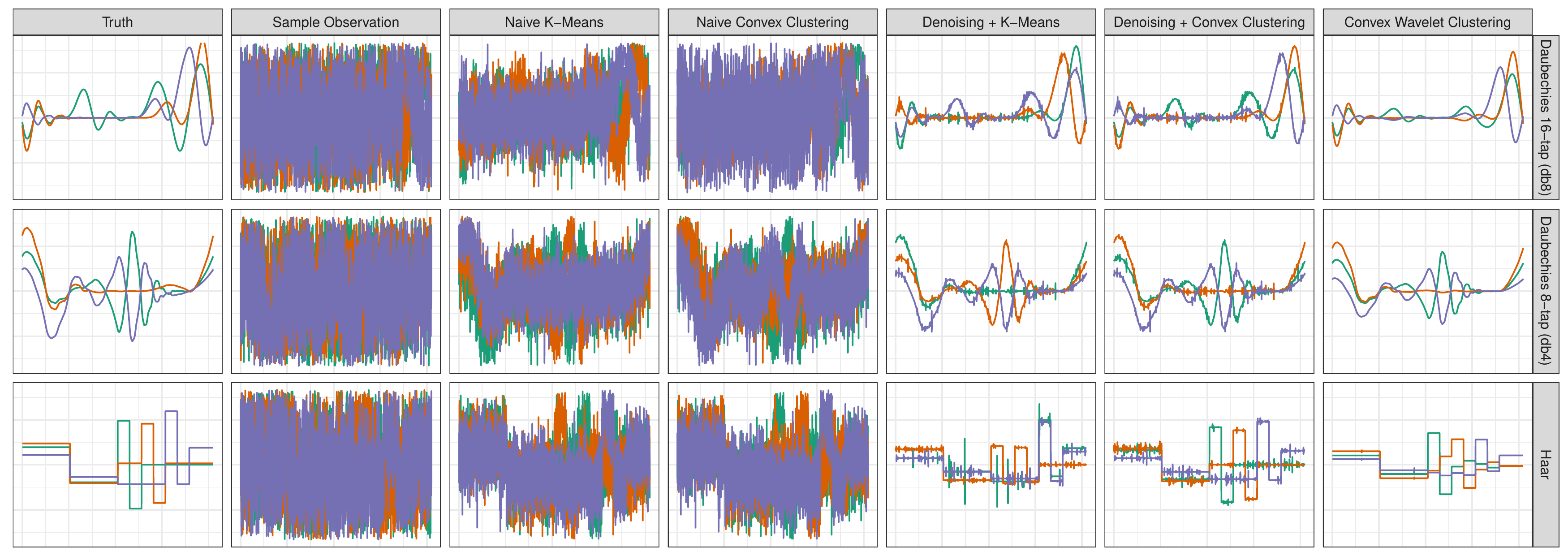}
\caption{
Sparse wavelet convex clustering on synthetic signals.
The top and middle rows use the Daubechies 16-tap (db8) and 8-tap (db4) filters, respectively, while the bottom row uses the Haar wavelet. From left to right: the baseline signals; a sample noisy observation (truncated for scale); centroids obtained from na\"ive $K$-means; na\"ive convex clustering; $K$-means applied to thresholded wavelet coefficients; convex clustering applied to thresholded wavelet coefficients; and our proposed method. In all three cases, our proposed method strikes a balance between clustering in a way that preserves salient features of the data, as well as admitting a simple, sparse representation.
}

\label{fig:signals}
\end{figure*}

\subsection{Background: Convex Clustering}
\citet{Pelckmans:2005} proposed a convex formulation of clustering, later popularized by \citet{Hocking:2011} and \citet{Lindsten:2011}. This formulation combines a Euclidean (Frobenius) loss function similar to that of $K$-means with a convex fusion penalty reminiscent of hierarchical clustering.  \emph{Convex clustering} is given as the solution to the following optimization problem, which clusters the rows of a data matrix $\bX \in \R^{n \times T}$:  
\begin{equation}
    \widehat{\bU}=\argmin_{\bU \in \R^{n \times T}} \frac{1}{2} \|\bU - \bX\|_F^2 + \lambda  \sum_{\substack{i, j = 1 \\ i < j}}^n w_{ij} \|\bU_{i \cdot} - \bU_{j\cdot}\|_q. \label{eqn:cvx_clust}
\end{equation}
Here, $\lambda \in \R_{\geq 0}$ is a regularization parameter which controls the degree of clustering induced in the matrix of centroids $\widehat{\bU} \in \R^{n \times T}$ and $\{w_{ij}\}$ are non-negative weights which incorporate prior information in the problem. Following \citet{Hocking:2011}, the unitarily invariant $\ell_2$-fusion penalty ($q = 2$) is typically used in practice. Two columns of $\bX$ are said to belong to the same cluster if the corresponding columns of $\widehat{\bU}$, parameterized by $\lambda$, are equal; that is, if they have the same estimated centroid. 
There has been much recent work developing algorithms for efficiently solving the convex clustering problem \citep{Chi:2015,Weylandt:2020,Panahi:2017,Sun:2021}.  

The basic convex clustering framework has been extended to induce additional structure in the estimated centroid matrix $\widehat{\bU}$ \citep{Chi:2017,Wang:2016-RobustCC,Chi:2018b}. Relevant to this paper, \citet{Wang:2018-SparseCC} add an $\ell_2$-penalty to the rows of $\bU$ to identify a sparse set of features which distinguish cluster centroids: 
\begin{equation}
    \frac{1}{2} \|\bU - \bX\|_F^2 + \lambda  \sum_{\substack{i, j = 1 \\ i < j}}^n w_{ij} \|\bU_{i\cdot} - \bU_{j\cdot}\|_q + \gamma \sum_{j = 1}^T \|\bU_{\cdot j}\|_2.\label{eqn:sparse_cvx_clust}
\end{equation}
For sufficiently large values of $\gamma \in \R_{\geq 0}$, the estimated cluster centroids differ on a small, sparse set of features. \citet{Wang:2018-SparseCC} motivate this approach in the high-dimensional setting, where many features are assumed to be pure noise. In this paper, we extend their approach to the wavelet domain and propose a novel method for simultaneous clustering and denoising by leveraging the fact that denoised cluster centroids should have sparse wavelet coefficients.  

\subsection{Background: Wavelet Denoising}

Many techniques in signal processing leverage \emph{transforms}, where representations in an alternate coordinate system shed light on the structure of signals not obvious from their time-domain representation.
One such type of transform is the \emph{wavelet transform}, where the signal of interest is projected onto a family of basis functions that localize information in the time and frequency domains.
Due to their spatial localization, coefficients in the wavelet domain capture transient features of time-domain signals at different scales.
Because of this behavior, natural signals with piecewise smooth structure admit sparse representations in the wavelet transform, motivating a variety of approaches for denoising~\cite{Donoho:1995b}, inpainting~\cite{Fadili:2009}, and compression~\citep{Taubman:2000}.
Indeed, there are immediate connections between wavelet denoising and feature selection using the lasso, as discussed by \citet{ZhaoOgden:2012, ZhaoChen:2015}.
The literature on wavelet analysis is extensive; for more complete coverage, we refer the reader to the excellent textbooks by \citet{Daubechies:1992} and \citet{Mallat:2009}, the review article by \citet{Antoniadis:2007}, and the draft monograph by \citet{Johnstone:2019}, as well as references therein.

\subsection{Background: Wavelet Clustering}

Our aim in this work is to incorporate sparsity in the wavelet domain to improve the performance of clustering algorithms. Wavelet denoising either before or after clustering has been studied by several authors; see, for instance, the framework developed by \citet{Misti:2007} or the many examples discussed by \citet{Aghabozorgi:2015}. These approaches almost exclusively proceed by calculating a wavelet representation, denoising via thresholding or feature selection, and applying a non-temporal clustering mechanism such as $K$-means to the denoised representation. \citet{Antoniadis:2013} give a readable overview of this framework, highlighting the effect of different denoising and clustering steps. Unified approaches, like that we propose below, are less common, though the approach of \citet{Ray:2006}, who combine a Dirichlet process prior with a wavelet representation of the cluster centroids in a Bayesian framework, allowing the user to incorporate prior information about the shape and regularity of cluster centroids, has similarities to our approach.

\subsection{Contributions}

Our contributions are as follows: we propose an extension of sparse convex clustering for application in the wavelet domain that jointly clusters and denoises signals.  In contrast to common practice which denoises and then clusters, we show that our approach yields wavelet-sparse centroids, aiding in interpretability and data compression.  Additionally, we develop an efficient ``Cartesian-Block'' ADMM algorithm for our problem and prove its linear convergence.  We also demonstrate the efficacy of sparse convex wavelet clustering through synthetic and real datasets, illustrating desirable properties compared to existing and commonly employed methods.

\section{Simultaneous Clustering and Wavelet Denoising}
We combine the sparse convex clustering approach of \citet{Wang:2018-SparseCC} with wavelet denoising techniques by minimizing the following optimization criterion:
\begin{equation}
    \frac{1}{2} \|\bU - \bX\|_F^2 + \lambda  \sum_{\substack{i, j = 1 \\ i < j}}^n w_{ij} \|\bU_{i\cdot} - \bU_{j\cdot}\|_2 + \gamma \sum_{j = 1}^T \omega_i \|\bU_{\cdot j}\bPsi\|_2, \label{eqn:wave_cvx_clust}
\end{equation}
where $\bPsi \in \R^{T \times T}$ is an orthogonal matrix encoding the discrete wavelet transform. Compared with traditional sparse convex clustering \eqref{eqn:sparse_cvx_clust}, the final term of \emph{wavelet sparse convex clustering} \eqref{eqn:wave_cvx_clust} selects only a small number of wavelet features by placing a group-lasso penalty on the wavelet coefficients $\bU\bPsi$.

The key feature of our approach is that it jointly performs clustering and denoising in a single (convex) optimization problem. Problem \eqref{eqn:wave_cvx_clust} inherits well-known theoretical advantages of convexity, including provable global optimality and robustness to noisy data, as well as the practical advantage of being able to jointly tune the fusion and denoising parameters ($\lambda, \gamma$). A closer examination reveals the key advantage of our approach: the estimated cluster centroids are wavelet-sparse \emph{by construction} due to the group-lasso penalty applied to $\bU\bPsi$.

To make this point more clear: we compare our approach to $K$-means clustering, either preceded by or followed by wavelet denoising. If wavelet denoising is performed before clustering, the estimated cluster centroids are no longer guaranteed to be sparse. Specifically, if a wavelet coefficient is thresholded at the noise level $\sigma$, approximately 32\% of coefficients will be non-zero (approximately 5\% of estimated coefficients will remain greater than $\sigma$ in absolute value even after denoising) and their mean will almost surely be non-zero. Denoising the results of $K$-means clustering can produce sparse solutions, but the quality of the initial clustering is significantly impaired by the undamped noise.

We note that most clustering methods, especially $K$-means, (Euclidean) hierarchical clustering, and convex clustering are unitarily invariant. In this setting, ``wavelet clustering'' without denoising, \emph{e.g.}, setting $\gamma=0$ in the convex wavelet clustering problem \eqref{eqn:wave_cvx_clust}, yields the same results as clustering directly in the time-domain.

\section{Algorithm and Tuning-Parameter Selection}
\begin{figure*}
  \centering
  \includegraphics[width=\textwidth]{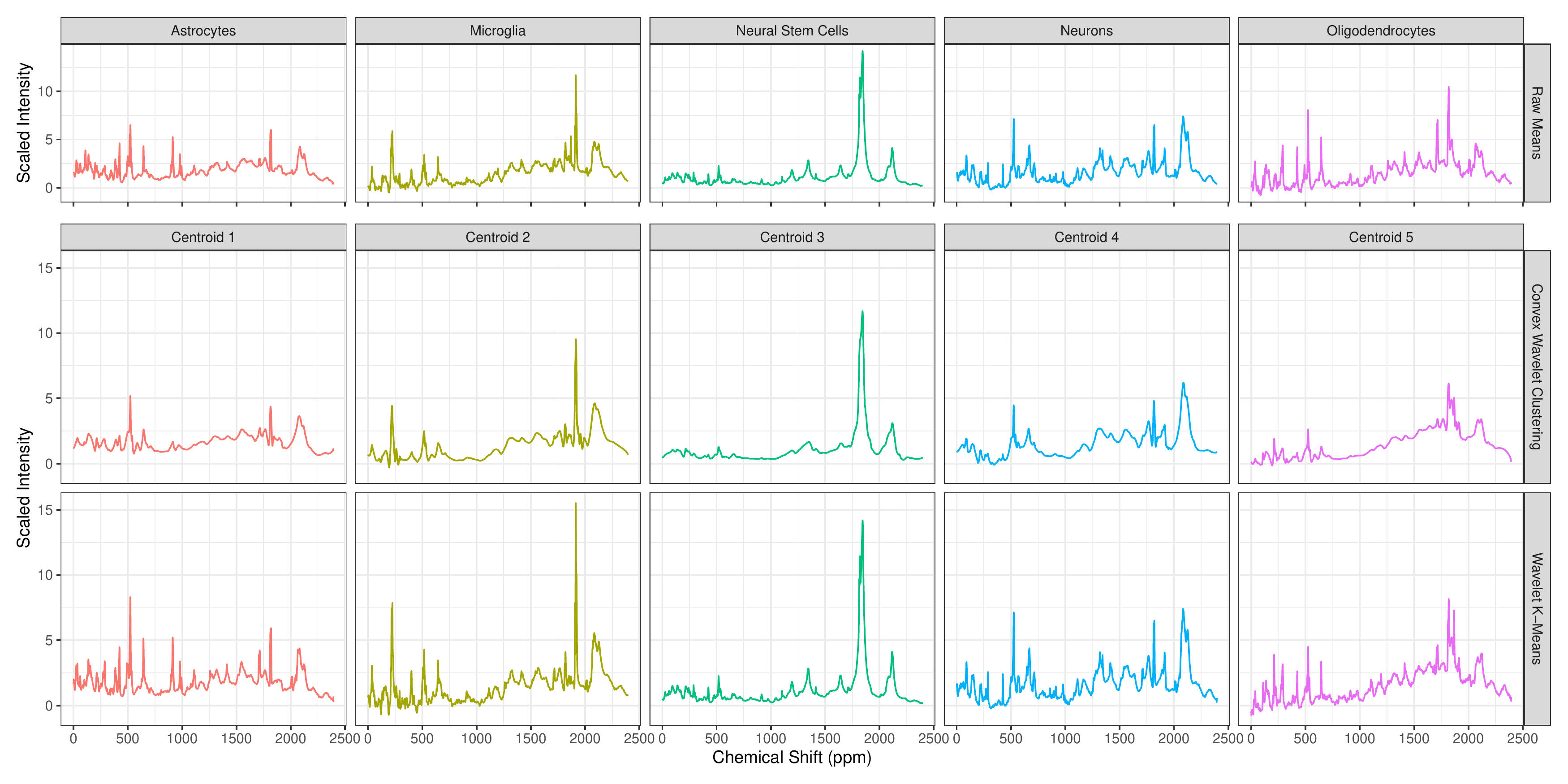}
  \caption{Results of the NMR Spectroscopy study discussed in Section \ref{sec:nmr}. The top row shows the sample means for each of the five known cell-types; the middle row shows the cluster centroids from wavelet denoising followed by $K$-means; and, the bottom row shows the results our sparse convex wavelet clustering method. Approaches used \texttt{db4} wavelets and used oracle tuning to fair comparisons.  Both methods are able to attain reasonable accuracy on this data set, as reflected by an Adjusted Rand Index of 66\%, but our approach yields more interpretable denoised and wavelet-sparse centroids, with clearly visible spikes that distinguish each cell type.}
  \label{fig:nmr}
\end{figure*}

Having defined our sparse wavelet convex clustering approach, we now turn to computational approaches for computing the solution of the sparse convex wavelet clustering problem \eqref{eqn:wave_cvx_clust} and selecting the fusion parameter ($\lambda$) and the denoising parameter ($\gamma$). In the case where $q = 2$, recalling that $\bPsi$ denotes an orthogonal transform, the convex wavelet clustering problem \eqref{eqn:wave_cvx_clust} is particularly easy to solve. Because both the Frobenius loss and the $\ell_2$ fusion penalty are invariant under orthogonal transformations, we can transform our signals to the wavelet domain, perform sparse convex clustering, and then apply the inverse transformation to the estimated centroid matrix. This approach allows us to take advantage of highly-efficient wavelet transforms \citep{Mallat:1989}, and is summarized in Algorithm \ref{alg:multi_admm}.

\begin{algorithm}
  \caption{Wavelet Sparse Convex Clustering Algorithm}\label{alg:multi_admm}
  \begin{itemize}
  \item \textbf{Input:}
  \begin{itemize}
      \item Data Matrix: $\bX \in \R^{n \times T}$
      \item Tuning Parameters: $\lambda, \gamma \in \R_{\geq 0}$
      \item Fusion Weights: $w_{ij} \in \R_{\geq 0}$
      \item Sparsity Weights: $\omega_{i} \in \R_{\geq 0}$
  \end{itemize}
  \item \textbf{Wavelet Transform:} $\bX^* = \bX\bPsi$
  \item \textbf{Perform Sparse Convex Clustering:} 
  \begin{itemize}
      \item $\tilde{\bU} = \text{Algorithm \ref{alg:scc_alg}}(\bX^*, \lambda, \gamma, \{w_{ij}\})$
  \end{itemize}
  \item \textbf{Back-Transform and Return}: $\hat{\bU} = \tilde{\bU}\bPsi^\top$
  \end{itemize}
\end{algorithm}
\noindent The core of Algorithm \ref{alg:multi_admm} is a sparse convex clustering problem. While the standard convex clustering problem is well-studied, existing approaches cannot be directly applied to the double penalty problem. \citet{Wang:2018-SparseCC} modify the ADMM approach of \citet{Chi:2015} and replace the primal update with a group lasso problem, which must be minimized by a secondary solver. Rather than using their approach, we propose a new ``Cartesian-Block'' ADMM, described in Algorithm \ref{alg:scc_alg}, for the sparse convex clustering problem. This approach does not require solving a group lasso problem and has closed-form updates for each step; in practice, this yields a significant performance boost.

\begin{algorithm}
  \caption{Cartesian-Block ADMM for Sparse Convex Clustering}\label{alg:scc_alg}
  \begin{itemize}
      \item \textbf{Input:}
      \begin{itemize}
      \item Data Matrix: $\bX \in \R^{n \times T}$
      \item Tuning Parameters: $\lambda, \gamma \in \R_{\geq 0}$
      \item Fusion Weights: $w_{ij} \in \R_{\geq 0}$
            \item Sparsity Weights: $\omega_{i} \in \R_{\geq 0}$
  \end{itemize}
      \item \textbf{Pre-Compute:} Directed Difference Matrix $\bD$
      \item \textbf{Initialize:} $\bV^{(0)} = \bZ^{(0)} = \bD\bX$
      \item \textbf{Repeat Until Convergence:}
\begin{align*}
    \bU^{(k+1)} &= \left[(1 + \rho)\bI + \rho\bD^\top\bD\right]^{-1}\\ &\quad\pp\left[\bX + \rho\bD(\bV_1^{(k)} - \bZ_1^{(k)}) + \rho(\bV_2^{(k)} - \bZ_2^{(k)})\right] \\
    \bV_1^{(k+1)} &= \textsf{prox}_{\lambda/\rho P_F(\cdot, \bw)}(\bD\bU^{(k+1)} + \bZ_1^{(k)}) \\
    \bV_2^{(k+1)} &= \textsf{prox}_{\gamma/\rho P_S(\cdot, \bomega)}(\bU^{(k+1)} + \bZ_2^{(k)}) \\
    \bZ_1^{(k+1)} &= \bZ_1^{(k)} + \bD\bU^{(k+1)} - \bV_1^{(k+1)} \\
    \bZ_2^{(k+1)} &= \bZ_2^{(k)} + \bU^{(k+1)} - \bV_2^{(k+1)}
\end{align*}
     \item \textbf{Return:} $\bU^{(k+1)}$
  \end{itemize}
\end{algorithm}

In Algorithm \ref{alg:scc_alg}, $\bD$ is a directed difference matrix corresponding to the differences between observations with non-zero fusion weights, $P_F(\cdot, \bw)$ is the $\bw$-weighted fusion-inducing column-wise group-lasso penalty, $P_S(\cdot, \bomega)$ is the $\bomega$-weighted sparsity-inducing row-wise group-lasso penalty, and $\prox_f(\cdot) = \argmin_{\bx} f(\bx) + \frac{1}{2}\|\bx - \cdot\|_F^2$ denotes the proximal operator. We defer the derivation of Algorithm \ref{alg:scc_alg} to the supplementary materials, but note that it can be considered a special case of the bi-clustering algorithm proposed by \citet{Weylandt:2019b}, with $\bD_{\text{row}} = \bD$ and $\bD_{\text{col}} = \bI$. Algorithm \ref{alg:scc_alg} has attractive convergence properties and exhibits linear convergence under relatively weak assumptions on $\bD$:

\begin{theorem}
Algorithm \ref{alg:scc_alg} exhibits primal, dual, and residual convergence for the sparse convex clustering problem \eqref{eqn:sparse_cvx_clust}. Furthermore, if $\bD$ has full row-rank, the convergence is linear. 
\end{theorem}

\noindent Convergence follows from standard ADMM convergence results, with the linear convergence result being a consequence of the strong convexity of the Frobenius loss and the rank assumptions on $\bD$ \citep{Hong:2017}. In situations where $\bD$ is not full-rank, a QR decomposition can be applied to find a full-rank matrix $\tilde{\bD}$ such $\bD\bU = \tilde{\bD}\bU$ for all $\bU$; replacing $\bD$ with $\tilde{\bD}$ in Problem \ref{eqn:sparse_cvx_clust} guarantees linear convergence while maintaining the same solution.

\begin{figure}[t]
    \centering
    \includegraphics[width=3.6in,height=2.4in]{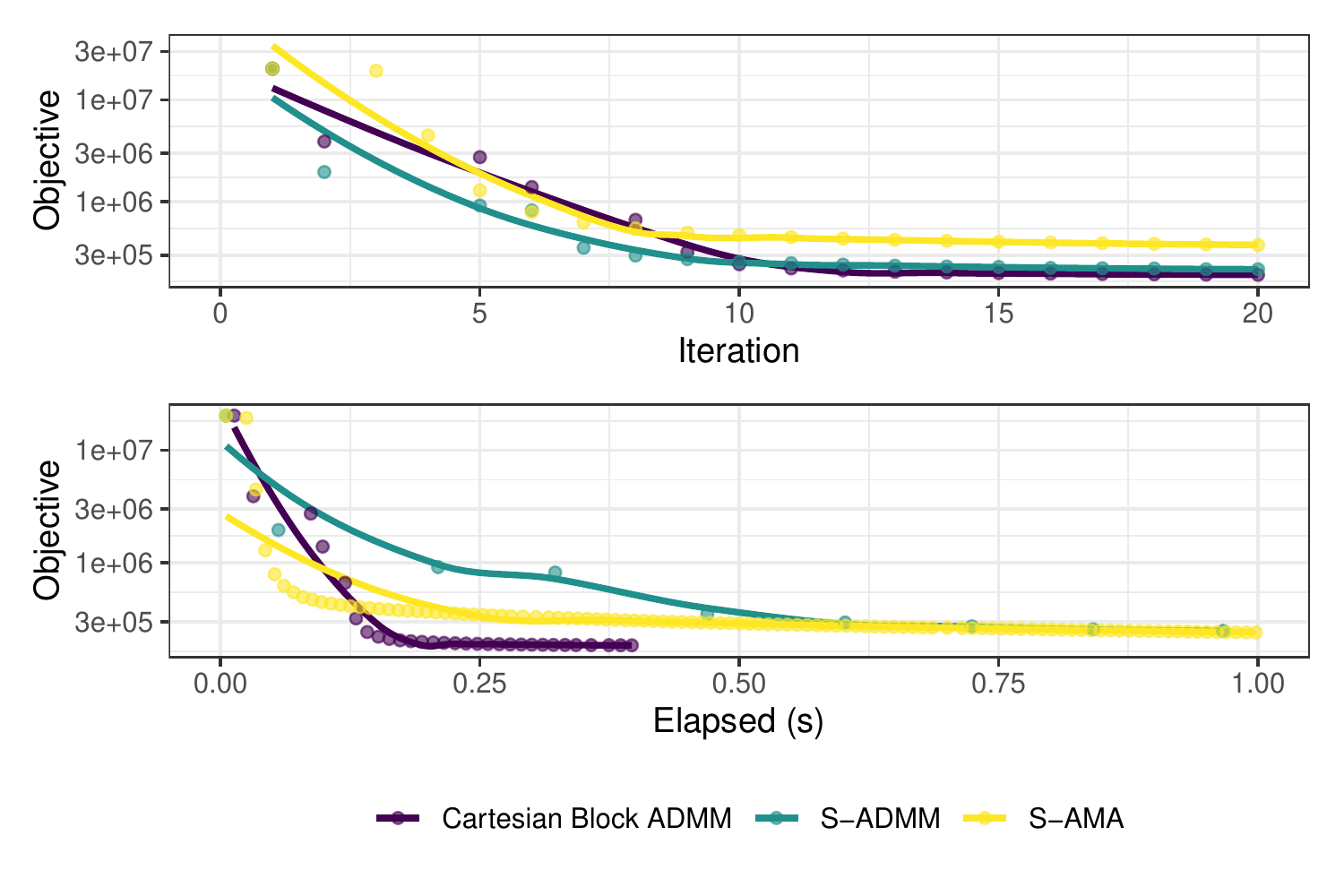}
    \caption{Timing comparison of Algorithm \ref{alg:scc_alg} with the \texttt{S-ADMM} and \texttt{S-AMA} algorithms of \citet{Wang:2018-SparseCC} on a simulated $\bX\in \R^{240 \times 1000}$ with three clusters and six non-noise features. The exact \text{S-ADMM} has the best \emph{per iteration} performance, but requires solving a group lasso problem at each iteration. Due to its simpler updates, Algorithm \ref{alg:scc_alg} has the best ``wall-clock'' performance.}
    \label{fig:timing}
\end{figure}

Computationally, Algorithm \ref{alg:scc_alg} significantly out-performs both the \texttt{S-ADMM} and \texttt{S-AMA} algorithms of \citet{Wang:2018-SparseCC}, as demonstrated in Figure \ref{fig:timing}. While the \emph{per iteration} performance of the \texttt{S-ADMM} is competitive with our method, \texttt{S-ADMM} requires a group-lasso problem to be solved at each iteration, significantly slowing its ``wall-clock'' performance. We note also that, for this problem, the \texttt{S-ADMM} is equivalent to the multi-block ADMM scheme suggested by \citet{Wang:2019-GECCO}. Unlike \citet{Wang:2019-GECCO}, who were only able to prove a relatively weak form of primal convergence, we establish primal, dual, and residual convergence generally, as well as a linear convergence rate under suitable $\bD$ and we do not require the use of a linearization (sub-problem approximation) scheme to achieve computational efficiency. While we do not discuss it in more detail here, this algorithm is also suitable for the one-step ``algorithmic regularization'' framework proposed by \citet{Weylandt:2020}, allowing for the entire clustering path (as a function of $\lambda$) to be efficiently recovered.

An important practical concern is how to select the fusion weights $\{w_{ij}\}$ and the sparsity weights $\{\omega_i\}$. We use the sparse Gaussian kernel weight scheme proposed by \citet{Chi:2015}, as implemented in the \texttt{clustRviz} \texttt{R} package for the fusion weights. 
In our experiments, we take inspiration from the empirical Bayes approach to wavelet denoising \citep{Johnstone:2005} and set $\omega_i = 1 - \zeta_j / \|\bzeta\|_1$ where $\zeta_j$ is the sample variance of the $j$-th wavelet coefficient.

\section{Experiments}

\subsection{Synthetic Data}

We demonstrate the efficacy of sparse convex wavelet clustering on synthetic signals. Figure~\ref{fig:signals} demonstrates the importance of \emph{simultaneous} as opposed to sequential denoising. For each wavelet basis (Haar, db4, db8), we consider three signals admitting a sparse representation in that basis, then add white noise (SNR of -7.7 dB) to each of five replicates, yielding $n=15$ total signals. We apply na\"ive $K$-means, na\"ive convex clustering \eqref{eqn:cvx_clust}, denoising followed by $K$-means, denoising followed by convex clustering, and our proposed method. The sequential denoising methods used the prescribed soft threshold of \citet{Donoho:1995b}. Compared to the other approaches, our method balances sparsity in the wavelet domain and structural fidelity. Qualitatively, sparse convex wavelet clustering yields centroids that are sparse by construction, unlike the other considered approaches.

More precisely, Table \ref{tab:sim_study} compares the performance of the considered approaches in terms of the adjusted Rand index (ARI) \citep{Hubert:1985}, correlation between the true and estimated centroids, compression (wavelet sparsity), and F1 score. Our method consistently out-performs the others in ARI, compression, and F1 score, while being out-performed by sequential denoising methods in correlation.  This could be explained by the additional bias of our approach, potentially correctable by debiasing (refitting) the inferred centroids after the fact.

\begin{table}[t]
    \centering
    \small
\begin{tabular}{crrrrr}
\toprule
Method & ARI & Correlation & Compression & F1 Score\\
\midrule
\multicolumn{5}{c}{Daubechies 16-Tap (db8)} \\
\midrule
KM & 90.91\% & 68.57\% & 0.39\% & 0.77\%\\
CC & 43.57\% & 60.48\% & 0.31\% & 0.61\%\\
D+KM & 81.83\% & 98.48\% & 93.95\% & 96.95\%\\
D+CC & {\bf 100\%} & {\bf 98.93\%} & 93.9\% & 96.94\%\\
CWC & {\bf 100\%} & 96.58\% & {\bf 99.62\%} & {\bf 99.93\%}\\
\midrule
\multicolumn{5}{c}{Daubechies 8-Tap (db4)} \\
\midrule
KM & 94.11\% & 69.11\% & 0.33\% & 0.66\%\\
CC & 63.91\% & 63.69\% & 0.31\% & 0.62\%\\
D+KM & 78.37\% & 98.05\% & 94.13\% & 97.02\%\\
D+CC & {\bf 100\%} & {\bf 99.53\%} & 94.09\% & 97.01\%\\
CWC & {\bf 100\%} & 99\% & {\bf 99.79\%} & {\bf 99.95\%}\\
\midrule
\multicolumn{5}{c}{Haar} \\
\midrule
KM & 84\% & 67.34\% & 0.36\% & 0.71\%\\
CC & 76.6\% & 65.13\% & 0.35\% & 0.69\%\\
D+KM & 82.34\% & 97.76\% & 94.12\% & 97.01\%\\
D+CC & {\bf 100\%} & {\bf 99.51\%} & 94.07\% & 97\%\\
CWC & {\bf 100\%} & 98.84\% & {\bf 99.71\%} & {\bf 99.9\%}\\
\bottomrule
\end{tabular}
  \caption{
  Performance of the na\"ive $K$-means (KM), na\"ive convex clustering (CC), wavelet denoising followed by $K$-means (D+KM), wavelet denoising followed by convex clustering (D+CC), and our proposed method (CWC). Each method is evaluated in terms of the adjusted rand index (ARI), correlation between the true and estimated centroids, wavelet sparsity (compression), and F1 score (a measure of support recovery in the wavelet domain). Our method consistently yields correct classification (ARI), and performs the best in recovering the true, sparse support in the wavelet domain (compression, F1 score). Notably, our approach is out-performed by the sequential denoising approaches (D+KM, D+CC), but still does better than the na\"ive approaches. Not shown here, clustering followed by denoising (KM+D, CC+D) inherits the low ARI of the na\"ive methods.
  }
  \label{tab:sim_study}
\end{table}

\subsection{Application to NMR Spectroscopy}\label{sec:nmr}

In this section, we apply our approach to a nuclear magnetic resonance (NMR) data set previously analyzed by \citet{Allen:2011} and compare it to standard wavelet clustering approaches. This data consists of the NMR spectra of 27 brain cells, discretized into bins of 0.04 parts per million (ppm), yielding 2394 different measurements of chemical shifts per sample. Five known cell types, collected as part of the original experiment, are used as cluster labels (astrocytes, $n = 4$; microglia, $n =9$; neural stem cells, $n = 7$; neurons, $n = 4$; oligodendrocytes, $n = 4$).  Each cell type is characterized by unique metabolites that resonate at particular chemical sifts, giving a cell-type signature.  But, due to the large amount of noise in this technology, these unique signatures are often obscured and the cell types are very difficult to distinguish (see the sample means for each cell type in the top portion of Figure~\ref{fig:nmr}).  

Before processing, all signals were normalized to have total power 1000.  We applied our approach and wavelet denoising followed by $K$-means using the  Daubechies wavelet with vanishing fourth moments (``db4''), with oracle threshold selection, though our results are quite robust to both the specific wavelet basis used and the threshold level. 

Na\"ive methods struggle with the high-dimensionality, small sample-size, and high-noise in this dataset.  Time-domain $K$-means achieves an ARI of just 45\% while standard convex clustering achieves an ARI of 63\%. Methods incorporating wavelet denoising perform better across all metrics. Both wavelet denoising followed by $K$-means and our approach achieve an ARI of 66\%. 

As with our examples shown above, the major advantage of sparse convex wavelet clustering is in the accuracy and the interpretability of the estimated centroids. Figure \ref{fig:nmr} compares the sample means from the known cell-types with those obtained by wavelet $K$-means and by our approach. (We manually aligned the estimated centroids and those for the known cell-types.) The centroids estimated by our method are highly wavelet-sparse (94.3\%), while the $K$-means centroids are only 28\% wavelet-sparse.  This sparse representation has a dual benefit: in addition to being highly compressible, it aids interpretation of the clustering results by highlighting the particular peaks that characterize each cell type. 

\section{Discussion}

We have proposed sparse convex wavelet clustering, a novel approach for simultaneous denoising and clustering of univariate signals based on a combination of wavelet denoising and sparse convex clustering. We have provided a provably-efficient algorithm for solving the  wavelet sparse convex clustering problem and demonstrated the effectiveness of our approach on synthetic and real NMR spectroscopy data. Compared to existing wavelet-based clustering techniques, ours combines denoising and clustering into a single step, rather than simply applying classical clustering techniques on a denoised wavelet representation. This simultaneous approach has several advantages: mathematically, it is unified and convex, providing global solutions to an otherwise challenging problem; practically, it yields centroids that are denoised (wavelet-sparse), aiding in interpretability of clustering results and data compression. Together, these advantages lead to improved clustering performance.

There are several possible extensions and further research related to our work.  First, our approach has two tuning parameters that control the clustering fusions and the amount of wavelet coefficient sparsity.  We have suggested possible data-driven tuning approaches in this paper, but further investigation may be warranted.  Related to this, there is an abundant literature on choosing the threshold for wavelet denoising, many with guarantees of theoretical optimality \citep{Donoho:1995b,Mallat:2009,Johnstone:2005}, which could inform the choice of the denoising parameter $\gamma$ and sparsity weights $\{\omega_j\}$. We also restricted our view to orthogonal transforms, but there is potential in using over-complete bases to impart stability and robustness \citep[Chapters 5 and 12]{Mallat:2009}. Finally, we use a simple penalty structure to induce sparsity in the wavelet coefficients, but, one could extend our approach to block-sparse penalties that reflect the hierarchical structure in the wavelet domain. Overall, we have proposed a new approach to simultaneous clustering and denoising that will find many applications and prompt many future investigations. 

\section{References}
{\ninept \printbibliography[heading=none]}
\end{refsection}

\begin{refsection}
\onecolumn
\appendix
\setcounter{algorithm}{0}
\renewcommand{\thealgorithm}{A\arabic{algorithm}}

\section*{Supplementary Materials}
\section{Derivations} \label{app:proofs}
\subsection{Derivation of Algorithm \ref{alg:multi_admm}}
In this section, we derive Algorithm \ref{alg:multi_admm} and show that a standard two-block ADMM in the wavelet domain can be used to solve the wavelet convex clustering problem \eqref{eqn:wave_cvx_clust}. For brevity, we elide the fusion weights $w_{ij}$ and sparsity weights $\omega_j$ and and introduce a directed difference matrix $\bD$ so that Problem \eqref{eqn:wave_cvx_clust} can be written as
\[\argmin_{\bU \in \R^{n \times T}} \frac{1}{2} \|\bU - \bX\|_F^2 + \lambda  \|\bU\bD\|_{q \rightarrow 1} + \gamma \|\bU\bPsi\|_{1 \leftarrow 2}\]
$\|\cdot\|_{q \rightarrow 1}$ denotes the sum of the row-wise $\ell_q$-norms, and
$\|\cdot\|_{1\leftarrow 2}$ denotes the sum of the column-wise $\ell_2$-norms. In the special
case where $\bPsi$ is orthogonal and $q = 2$, 
\[\argmin_{\bU \in \R^{n \times T}} \frac{1}{2} \|(\bU - \bX)\bPsi\|_F^2 + \lambda  \|\bU\bD\bPsi\|_{2 \rightarrow 1} + \gamma \|\bU\bPsi\|_{1\leftarrow 2}\]
We note that $\bD$ and $\bPsi$ can be swapped because they are inside an $\ell_2$ norm, so this becomes: 
\[\argmin_{\bU \in \R^{n \times T}} \frac{1}{2} \|(\bU\bPsi) - (\bX\bPsi)\|_F^2 + \lambda  \|(\bU\bPsi)\bD\|_{2 \rightarrow 1} + \gamma \|\bU\bPsi\|_{1\leftarrow 2}\]
Letting $\bU^* = \bU\bPsi$ and $\bX^* = \bX\bPsi$, it suffices to solve
\[\argmin_{\bU^* \in \R^{n \times T}} \frac{1}{2} \|\bU^* - \bX^*\|_F^2 + \lambda  \|\bU^*\bD\|_{2\rightarrow 1} + \gamma \|\bU\bPsi\|_{1 \leftarrow 2}\]
and then post-multiply our solution by $\bPsi^\top$. To solve this inner problem, we adapt the ``Hilbert Lifting ADMM'' trick presented by \citet{Weylandt:2019b} for the convex bi-clustering problem. To reduce clutter, we omit the star superscripts as we derive our algorithm. 

Let $\mathfrak{L}_1(\bU) = (\bD\bU, \bU)$ and let $\mathfrak{L}_2([\bV_1, \bV_2]) = [-\bV_1, -\bV_2]$ be the negative identity transform. The above problem can then be written as 
\begin{align*}
  \argmin_{\bU, \bV_1, \bV_2} &\pp \frac{1}{2} \|\bU - \bX\|_F^2 + \lambda  \|\bV_1\|_{2 \rightarrow 1} + \gamma \|\bV_2\|_{1 \leftarrow 2} \\
  \text{subject to  } &\pp \mathfrak{L}_1(\bU) - \mathfrak{L}_2([\bV_1, \bV_2]) = \bzero
\end{align*}
The (scaled) augmented Lagrangian for this problem is
\[\mathscr{L} = \frac{1}{2}\|\bU - \bX\|_F^2 + \lambda\|\bV_1\|_{2 \rightarrow 1} + \gamma \|\bV_2\|_{1 \leftarrow 2} + \frac{\rho}{2}\left\|(\bD\bU, \bU) - (\bV_1, \bV_2) + (\bZ_1, \bZ_2)\right\|^2\]
Before deriving the ADMM iterates, we note that this factorizes as 
\[\mathscr{L} = \frac{1}{2}\|\bU - \bX\|_F^2 + \lambda\|\bV_1\|_{2\rightarrow 1} + \frac{\rho}{2}\left\|\bD\bU - \bV_1 + \bZ_1\right\|_F^2 + \gamma \|\bV_2\|_{1\leftarrow 2} + \frac{\rho}{2}\left\|\bU - \bV_2 + \bZ_2\right\|^2\]
In this form, the primal ($\bU$) update is given by: 
\begin{align*}
  \argmin_{\bU} \mathscr{L} &= \argmin_{\bU} \frac{1}{2}\|\bU - \bX\|_F^2 + \frac{\rho}{2}\left\|\bD\bU - \bV_1 + \bZ_1\right\|_F^2 + \frac{\rho}{2}\left\|\bU - \bV_2 + \bZ_2\right\|^2
\end{align*}
Differentiating with respect to $\bU$, we see that the stationary conditions of the $\bU$-subproblem are
\[\bzero = (\bU - \bX) + \rho\bD^\top(\bD\bU - \bV_1 + \bZ_1) + (\bU - \bV_2 + \bZ_2)\]
which has the analytical solution: 
\[\bU = \left[(1 + \rho)\bI + \rho\bD^\top\bD\right]^{-1}\left[\bX + \rho\bD^\top(\bV_1 - \bZ_1) + \rho(\bV_2 - \bZ_2)\right]\]

The copy update for $\bV_1$ is given by:
\begin{align*}
  \argmin_{\bV_1} \mathscr{L} &= \argmin_{\bV_1} \lambda\|\bV_1\|_{2\rightarrow 1} + \frac{\rho}{2}\left\|\bD\bU - \bV_1 + \bZ_1\right\|_F^2 = \textsf{prox}_{\lambda/\rho \|\cdot\|_{2\rightarrow 1}}(\bD\bU + \bZ_1)
\end{align*}
and similarly for $\bV_2$: 
\begin{align*}
  \argmin_{\bV_2} \mathscr{L} &= \argmin_{\bV_2} \gamma \|\bV_2\|_{1\leftarrow 2} + \frac{\rho}{2}\left\|\bU - \bV_2 + \bZ_2\right\|_F^2 = \textsf{prox}_{\gamma / \rho \|\cdot\|_{1\leftarrow 2}}(\bU + \bZ_2)
\end{align*}
Hence, the combined ADMM iterates are for the sparse convex clustering problem are: 
\begin{align*}
    \bU^{(k+1)} &= \left[(1 + \rho)\bI_{n \times n} + \rho\bD^\top\bD\right]^{-1}\left[\bX + \rho\bD^\top(\bV_1^{(k)} - \bZ_1^{(k)}) + \rho(\bV_2^{(k)} - \bZ_2^{(k)})\right] \\
    \bV_1^{(k+1)} &= \textsf{prox}_{\lambda \|\cdot\|_{2 \rightarrow 1}}(\bD\bU^{(k+1)} + \bZ_1^{(k)}) \\
    \bV_2^{(k+1)} &= \textsf{prox}_{\gamma \|\cdot\|_{1 \rightarrow 2}}(\bU^{(k+1)} + \bZ_2^{(k)}) \\
    \bZ_1^{(k+1)} &= \bZ_1^{(k)} + \bD\bU^{(k+1)} - \bV_1^{(k+1)} \\
    \bZ_2^{(k+1)} &= \bZ_2^{(k)} + \bU^{(k+1)} - \bV_2^{(k+1)}
\end{align*}
This finishes the derivation of Algorithm \ref{alg:scc_alg}. Combining these updates with the wavelet discussion above yields Algorithm \ref{alg:multi_admm}. Note that the $\bV$ and $\bZ$ updates can each be computed in parallel. In practice, the Cholesky factorization of $(1 + \rho)\bI_{n \times n} + \rho \bD^\top\bD$ can be cached and re-used between iterations.

For comparison, we restate other methods proposed for the sparse convex clustering problem in our notation. The \texttt{S-ADMM} of \citet{Wang:2018-SparseCC} consists of the following iterates:
\begin{align*}
\bU^{(k+1)} &= \argmin_{\bU} \frac{1}{2}\|\begin{pmatrix} \bI \\ \sqrt{\rho}\bD \end{pmatrix}\bU - \begin{pmatrix} \bX \\ \rho^{1/2}(\bV^{(k)} - \bZ^{(k)})\end{pmatrix}\|_F^2 + \gamma\|\bU\|_{1 \leftarrow 2} \\
            &= \textsf{Multi-Group-Lasso}\left(\tilde{\bX} = \begin{pmatrix} \bI \\ \sqrt{\rho}\bD \end{pmatrix}, \tilde{\bY} = \begin{pmatrix} \bX \\ \rho^{1/2}(\bV^{(k)} - \bZ^{(k)})\end{pmatrix}\gamma\right) \\
\bV^{(k+1)} &= \prox_{\lambda/\rho\|\cdot\|_{2 \rightarrow 1}}\left(\bD\bU^{(k+1)} + \bZ^{(k)}\right) \\
\bZ^{(k+1)} &= \bZ^{(k)} + \bD\bU^{(k+1)} - \bV^{(k+1)}
\end{align*}
where $\textsf{Multi-Group-Lasso}$ is a call to a secondary solver for a multi-task regression problem with group lasso penalty, for which several efficient algorithms are available \citep{Obozinski:2011}.

The \texttt{S-AMA} of \citet{Wang:2018-SparseCC} consists of the following iterates:
\begin{align*}
\bU^{(k+1)} &= \prox_{\gamma \|\cdot\|_{1 \leftarrow 2}}(\bX - \bD^{\top}\bZ) \\
\bV^{(k+1)} &= \prox_{\lambda/\rho\|\cdot\|_{2 \rightarrow 1}}\left(\bD\bU^{(k+1)} + \rho^{-1}\bZ^{(k)}\right) \\
\bZ^{(k+1)} &= \bZ^{(k)} + \rho(\bD\bU^{(k+1)} - \bV^{(k+1)})
\end{align*}
Note that the dual variable $\bZ$ is a scaled version of that used for the various ADMM iterates. The AMA is able to omit the quadratic penalty term in the augmented Lagrangian in the $\bU$ update and hence avoid having to solve a full linear system. 

Further efficiency gains in the \texttt{S-AMA} can be simplified using Moreau's decomposition \citep{Moreau:1962} to elide the $\bV$-variable. In particular, Moreau's result\footnote{Restricted to $f(\cdot) = \lambda \|\cdot\|$ for some $\lambda \in \R_{\geq 0}$ and some norm, Moreau's identity implies
\[\bx = \prox_f(\bx) + \Pi_{\mathcal{B}^*(\lambda)}(\bx)\] where $\mathcal{B}^*$ is the dual norm ball of radius $\lambda^{-1}$. In our case, we use the relationship: 
\[\bx - \prox_f(\bx) = \Pi_{\mathcal{B}^*(\lambda)}(\bx)\]
See also Section 2.5 of the monograph by \citet{Parikh:2014}} allows us
to re-write the copy and dual updates as: 
\begin{align*}
\bZ^{(k+1)} &= \bZ^{(k)} + \rho(\bD\bU^{(k+1)} - \bV^{(k+1)}) \\ 
            &= \bZ^{(k)} + \rho\left(\bD\bU^{(k+1)} - \prox_{\lambda/\rho \|\cdot\|_{2\rightarrow 1}}(\bD\bU^{(k+1)} + \rho^{-1}\bZ^{(k)})\right) \\ 
\rho^{-1}\bZ^{(k+1)} &= \rho^{-1}\bZ^{(k+1)} + \bD\bU^{(k+1)} - \prox_{\lambda/\rho \|\cdot\|_{2\rightarrow 1}}(\bD\bU^{(k+1)} + \rho^{-1}\bZ^{(k)}) \\
                     &= \Pi_{\lambda/\rho \mathcal{B}_{\|\cdot\|_{2\rightarrow 1}^*}}(\bD\bU^{(k+1)} + \rho^{-1}\bZ^{(k)})
\end{align*}
where the projection is onto the dual ball of the $\|\cdot\|_{2\rightarrow 1}$-norm with radius $\lambda / \rho$. The combined updates are thus: 
\begin{align*}
\bU^{(k+1)} &= \prox_{\gamma \|\cdot\|_{1 \leftarrow 2}}(\bX - \bD^{\top}\bZ) \\
\bZ^{(k+1)} &= \rho \Pi_{\lambda/\rho \mathcal{B}_{\|\cdot\|_{2\rightarrow 1}^*}}(\bD\bU^{(k+1)} + \rho^{-1}\bZ^{(k)})
\end{align*}
where $\Pi_{\lambda/\rho\mathcal{B}_{\|\cdot\|_{2\rightarrow1}^*}}(\cdot)$ denotes projection onto the dual ball of the $\|\cdot\|_{2\rightarrow 1}$-norm with radius $\lambda / \rho$. This simplicity comes at a cost however: while the ADMM converges for any $\rho$, or indeed even variable $\rho$, the AMA imposes a step-size bound on $\rho$, depending on the strong convexity of the objective. Since the additional sparse penalty term does not add strong convexity, we can use the same bound
\[\rho < \frac{2}{\lambda_{\max}(\bD^{\top}\bD)}\]
See \citet[Section 4.2]{Chi:2015} \citet[Appendix A]{Weylandt:2019b} for derivation of this bound. This smaller step-size typically significantly limits per iteration performance of the \texttt{S-AMA}.

\citet{Wang:2019-GECCO} propose a form of multi-block ADMM to solve
integrative sparse generalized convex clustering. Specializing their approach, as 
given in their Algorithm 2, to one Gaussian data view, we see that it consists of the ADMM updates: 
\begin{align*}
\bU^{(k+1)} &= \argmin_{\bU} \frac{1}{2}\|\bU - \bX\|_F^2 + \frac{\rho}{2}\|\bD\bU - \bV^{(k)} + \bZ^{(k)}\|_F^2 + \gamma \|\bU\|_{1 \leftarrow 2}\\
\bV^{(k+1)} &= \prox_{\lambda / \rho \|\cdot\|_{2\rightarrow 1}}(\bD\bU^{(k+1)} + \bZ^{(k)}) \\ 
\bZ^{(k+1)} &= \bZ^{(k)} + \bD\bU^{(k)} - \bV^{(k)}
\end{align*}
To solve the $\bU$-subproblem, we note that this is the same $\bU$-update used in the \texttt{S-ADMM} above, 
and hence can be solved as a multivariate group-lasso problem: 
\begin{align*}
\bU^{(k+1)} &= \argmin_{\bU} \frac{1}{2}\|\bU - \bX\|_F^2 + \frac{\rho}{2}\|\bD\bU - \bV^{(k)} + \bZ^{(k)}\|_F^2 + \gamma \|\bU\|_{1 \leftarrow 2}\\
            & \textsf{Multi-Group-Lasso}\left(\tilde{\bX} = \begin{pmatrix} \bI \\ \sqrt{\rho}\bD \end{pmatrix}, \tilde{\bY} = \begin{pmatrix} \bX \\ \rho^{1/2}(\bV^{(k)} - \bZ^{(k)})\end{pmatrix}; \gamma\right) \\
\bV^{(k+1)} &= \prox_{\lambda / \rho \|\cdot\|_{2\rightarrow 1}}(\bD\bU^{(k+1)} + \bZ^{(k)}) \\ 
\bZ^{(k+1)} &= \bZ^{(k)} + \bD\bU^{(k)} - \bV^{(k)}
\end{align*}

As \citet{Wang:2019-GECCO} note, it is not necessary to solve the $\bU$-problem of the ADMM to completion. They demonstrate that taking only a single proximal gradient step suffices to establish convergence and often significantly out-performs fully solving the primal problem. In our notation, their Algorithm 5 becomes: 
\begin{align*}
\bU^{(k+1)} &= \prox_{\gamma\|\cdot\|_{1\leftarrow 2}}\left((1 - s)\bU^{(k)} + s\bX - s\rho\bD^{\top}(\bD\bU^{(k)} - \bV^{(k)} + \bZ^{(k)})\right)\\
\bV^{(k+1)} &= \prox_{\lambda / \rho \|\cdot\|_{2\rightarrow 1}}(\bD\bU^{(k+1)} + \bZ^{(k)}) \\ 
\bZ^{(k+1)} &= \bZ^{(k)} + \bD\bU^{(k)} - \bV^{(k)}
\end{align*}
for $s = 1/\lambda_{\max}(\bI + \rho \bD^{\top}\bD)$. 

To obtain this, this note that we have a single loss function, so their $k$ (block index) is
equal to one throughout, as is $\pi_k$. Then specialize the primal ($\bU$) update (their Algorithm 3) with $\ell(\bU, \bX) = \frac{1}{2}\|\bU - \bX\|_F^2$ which has $\nabla \ell = \bU - \bX$ and $\nabla^2\ell = \bI$; combining
this with the augmented Lagrangian, we have $\nabla^2_{\text{Smooth Terms}} = \bI + \rho\bD^{\top}\bD$ and we fix $s^{-1} = \lambda_{\max}(\bI + \rho\bD^{\top}\bD)$. Substituting this into the proximal gradient
update (their Algorithm 3), we find
\begin{align*}
\bU^{(k+1)} &= \prox_{\gamma\|\cdot\|_{1 \leftarrow 2}}\left(\bU^{(k)} - s\left[\bU^{(k)} - \bX + \rho\bD^{\top}(\bD\bU^{(k)} - \bV^{(k)} + \bZ^{(k)})\right]\right) \\
&= \prox_{\gamma\|\cdot\|_{1\leftarrow 2}}\left((1 - s)\bU^{(k)} + s\bX - s\rho\bD^{\top}(\bD\bU^{(k)} - \bV^{(k)} + \bZ^{(k)})\right)
\end{align*}
Interestingly, this approach seems to lie somewhere between the AMA and the
Cartesian block ADMM algorithms.

\section{Additional References}
\printbibliography[heading=none]

\end{refsection}
\end{document}